\pdfoutput=1

\documentclass[11pt]{article}

\newcommand{\cready}[1]{}

\usepackage[preprint]{acl}

\usepackage{times}
\usepackage{latexsym}
\usepackage{booktabs}
\usepackage{amsmath}

\usepackage[T1]{fontenc}

\usepackage[utf8]{inputenc}

\usepackage{microtype}

\usepackage{inconsolata}

\usepackage{graphicx,caption, }
\usepackage{multirow}
\usepackage{subcaption}
\title{Pretraining Language Models for Diachronic Linguistic Change Discovery}

\author{
    Elisabeth Fittschen\textsuperscript{1}, 
    Sabrina Li\textsuperscript{2},
    Tom Lippincott\textsuperscript{2},
    Leshem Choshen\textsuperscript{3},
    Craig Messner\textsuperscript{2}\\[1em]
    \textsuperscript{1}University of Hamburg, Germany\\
    \textsuperscript{2}Center for Digital Humanities, Johns Hopkins University, USA\\
    \textsuperscript{3}IBM Research, MIT, USA\\
    \\
}

\begin{document}

\maketitle

\begin{abstract}
Large language models (LLMs) are increasingly used as knowledge discovery tools. Humanistic disciplines like historical linguistics and literary studies have shown interest in this capability. These fields often construct arguments on the basis of distinctions between phenomena like time-period or genre. Such methodological investments complicate reliance on LLMs pretrained over large sets of broadly-collected data. We show that efficient pretraining techniques produce useful models of semantic change over modest historical corpora without allowing potential contamination from anachronistic data. We verify that these trained-from-scratch models better respect historical divisions and are more computationally efficient compared to the standard approach of fine-tuning an existing LLM.  We compare the trade-offs in general linguistic fluency versus detecting and characterizing various forms of linguistic change, and provide a pipeline implementation of our approach that can be readily adapted and applied to a wide range of diachronic phenomena.
\end{abstract}

\section{Introduction}
Certain fields of research invest heavily in the importance of boundaries that demarcate their objects of study into groups. These distinctions might include general phenomena like the arrow of time (e.g. diachronic change in linguistics) or specific boundaries drawn from traditional means of practice (e.g. forms of poetry in literary studies).

Such methodological investments are somewhat at odds with the dominant modern technology for language research, pretraining large language models (LLMs) \citep{underwood2025can}. LLMs are at least in part successful due to their omnivorous nature \citep{kaplan2020scaling}; they develop general skills by consuming as diverse and as large a corpus as possible \citep{polo2024sloth}. In contrast, our target fields are characterized by both limited data and particular interests. For example, diachronic linguistics cares more about time-specific aspects of language rather than general model capability.
LLMs do have some ability to divide information (for example, produce a haiku and not a limerick in a zero-shot setting) \citep{cheng2024dated,ifergan2024beneath}. However, prompting or other elicitation techniques may not offer immediate evidence that a particular output relied only on knowledge appropriate to a  period or division. We propose a straightforward solution. To ensure a model's weights contain no proscribed information, you can train a model on a restricted corpus of your own choosing.

We show that pretraining under a tight academic budget of data (and compute) proves surprisingly effective when performed using the efficient methods developed by the BabyLM community \citep{conll-2024-babylm}. Although those techniques are designed for data efficiency and cognitively-plausible pretraining, we find that BabyLlama-2 \citep{tastet-timiryasov-2024-babyllama} is also a useful recipe for academic pretraining.

We leverage this form of efficiency by developing a multiple-model approach that offers exploratory access to corpus-level hypotheses concerning lexical change, non-lexical (grammatical and morphological) change, and word sense introduction/obsolescence.
Specifically, we train a series of 5 models each with a pretraining dataset of 10 million tokens drawn from consecutive historical periods.
We evaluate these pretrained models (\emph{scratch}), as well as larger models finetuned on the same data slices (\emph{finetuned}), using standard metrics and evaluation sets. We also evaluate both model sets along the axis of historical specificity by determining how much information "leakage" between periods each approach allows.  Finally, we demonstrate our scratch method's ability to analyze diachronic linguistic change using a novel cloze evaluation set, qualitative analysis and use-case examples. 
We find that:
\begin{enumerate}
    \item The scratch models train nearly two times faster than DoRA finetuned models while retaining adequate performance for many tasks
    \item The finetuned models "leak" information across time periods in a way that our models do not, justifying a tradeoff of test set performance for specificity
    \item The full battery of scratch models can be used to generate hypotheses about linguistic change across our corpus
\end{enumerate}

These qualities render our approach useful for hypothesis discovery and exploration in fields that demand firm lines of epistemological demarcation.

\section{Related work}
\subsection{LLM analysis of lexical semantic change}
Multiple works suggest that LLMs could serve linguistic studies in flexible ways not imagined before \citep{warstadt2022artificial}, such as simulating human subject responses \citep{Wilcox2020OnTPA,Trott2024CanLL,Aher2022UsingLLA} or modeling language acquisition \citep{conll-2024-babylm,warstadt-etal-2023-findings}. We draw upon this thinking, calling for pretraining as a now feasible and promising way of contrasting corpora and aiding linguistic queries, broadly, and specifically for this in diachronic lexical semantic change. Lexical semantic change is a field dedicated to finding words that changed in meaning. Work in this field have employed LLMs, initially embeddings from masked language models (MLMs) \citep{periti2024lexical}. This approach leads to issues processing and aligning embeddings \citep{schlechtweg-etal-2020-semeval} and potentially complicates change discovery outside of benchmark environments \citep{umarova-etal-2025-current}. Causal modeling has also been employed for diachronic change detection. However, prior approaches largely use causal models to modify \citep{periti-etal-2024-analyzing} or augment \citep{periti-etal-2024-automatically} text samples before processing them with downstream MLMs. Other work has used inference from pretrained autoregressive models to directly measure change, relying on in-context learning and RAG to guide the model's judgments \citep{periti-etal-2024-chat, hagen-2025-lexical}. While these approaches have shown potential, our approach of pretraining contrasting efficient models both guarantees division between domains and offers the possibility that the models capture features that extend past lexical meaning.

\subsection{Efficient finetuning}
Increased finetuning costs have lead to the development of parameter-efficient rank-adapter finetuning techniques like LoRA \citep{hu2022lora} and DoRA \citep{liu2024dora}. As these techniques are the most similar in compute and token demands to our approach we use DoRA finetuning to produce comparison performance and specificity scores. We select DoRA over LoRA due to its generally reported higher performance.

\subsection{Domain-specific language modeling}
Previous embedding models approaches to domain specific modeling have often chosen BERT-like architectures, employing finetuning \citep{hosseini2021neural, qiu2022histbert} and pretraining \citep{beck2023ghisbert, manjavacas2022adapting}. However, beyond being non-causal, these strategies employ large sets of data, limiting their ability to be adapted for smaller domains of interest. One project trains a recent-history-aware (2011 to 2022) model on GPT2, but does so in order to detect knowledge-level analogies rather than provide a methodological lever, and also employs a large dataset \citep{drinkall-etal-2024-time}.

\subsection{Evaluating and guaranteeing historically-specific model knowledge} Datasets for evaluating diachronic model knowledge have previously focused on historical performance, both linguistic \citep{manjavacas2022adapting} and at the level of knowledge \citep{dhingra2022time, piryani2024chroniclingamericaqa}
, with these latter sets typically being structured in QA form. 

\begin{table*}[h]
    \centering
    \begin{tabular}{p{11cm} c c}
        \toprule
        \textbf{Text} & \textbf{Sense Year} \\
        \midrule
        ``They had a bunch of crazy ideas that would never \textbf{work}'' & 1599 \\
        ``I tried to call the operator but the phone was \textbf{dead}'' & 1882 \\
        ``You know how it is. I'm not into ironing. It's not my \textbf{thing}'' & 1936 \\
        ``Let's go where there's some life. Whatta ya say? Hey baby, I'm \textbf{down}'' & 1952 \\
        \bottomrule
    \end{tabular}
    \caption{Cloze task examples and the year when the word sense first appeared}
    \label{tab:word_sense}
\end{table*}

\section{Experiments}
\subsection{Setup}
\textbf{Training data.} We employ a multistage pipeline to prepare time-period specific training data. This pipeline integrates three sources to estimate the publication date of each document found in the Project Gutenberg collection.
(1) Author information sourced from WikiData (\emph{WD}) (2) Work metadata found in the Project Gutenberg Catalog (\emph{PGC}) and (3) Inference performed by \emph{LLMs}

We define our overall historical range as the years 1750-1940, inclusive. We use a fuzzy string matching system, described in Appendix~\ref{sec:wdad} to align authors from WD to PGC works. We acquire publication dates for each author-aligned work by prompting an instruction-tuned LLM. We evaluate this work-date attribution approach over a variety of open- and closed-source LLMs by calculating their performance against a gold-annotated test set consisting of works published from 1550-1850. While closed-source LLMs perform best, Llama3.3-70B \citep{grattafiori2024llama}, quantized to 4 bits using the BitsAndBytes library \cite{belkada2023making} performs well enough to justify its use (see Appendix~\ref{sec:wdad} for evaluation details).

We use this information to split the full corpus into 5 diachronic slices. We set the boundaries for these slices by negotiating between minimizing their duration and obtaining 10 million training tokens for each split. We further reserve 5 million tokens from each split for testing and 1 million for validation during training. This results in 5 equal subcorpora for the time periods 1750-1820, 1820-1850, 1850-1880, 1880-1910, and 1910-1940.
\subsection{Procedure}
\textbf{Model training.} 
We employ two primary training approaches over each split of the historical data: (1) \emph{finetuned} models adapted from a larger pretrained model, and (2) \emph{scratch} models trained solely on the historical datasets. We train 3 full batteries of scratch models with unique random initializations and training set shuffles. We present all scratch results with a 1 standard deviation uncertainty region calculated from these runs.

We train the finetuned models using DoRA adapters on top of a Llama3-8B backbone. We train the scratch models using the BabyLlama-2 recipe, which employs a distillation approach. Ultimately, training the scratch BabyLlama-2 models proved quicker and more efficient than the DoRA finetuning process (see Appendix~\ref{sec:adddetails}). 

As an additional point of comparison, we also produce a battery of models derived from the released checkpoint of BabyLlama-2. We DoRA finetune this released checkpoint, which was pretrained over the BabyLM data mixture, with each slice of our historical data, resulting in the \emph{small} battery of models. This small set underperformed the scratch and finetuned approaches, and was consequently set aside (see Appendix~\ref{sec:smallperf}) \footnote{Models and datasets are found \href{https://huggingface.co/Hplm}{here}, the code repository is found \href{https://github.com/comp-int-hum/historical-perspectival-lm}{here}.}

\section{Evaluation}
We evaluate the scratch and finetuned models using perplexity, a modified version of the BabyLM evaluation pipeline, and a novel cloze-structured evaluation set. Additionally, when appropriate, we contextualize the scratch evaluations using the released checkpoint of BabyLlama-2 (\emph{unadapted} BabyLlama-2) to calibrate expectations of typical performance of data-efficient models. Similarly, when appropriate, we contextualize the finetuned evaluations using distribution Llama-3-8B as a baseline (\emph{unadapted} Llama).

We use all three of these evaluation approaches to characterize a given model's general \textbf{fluency} and \textbf{historical specificity}. We further employ the cloze set to demonstrate the scratch approach's \textbf{exploratory potential.}

\begin{table*}[h]
    \centering
    \begin{tabular}{l l}
        \toprule
          \textbf{Model}  & \textbf{Sentence} \\
          \midrule
 1750 to 1820 &  \textcolor[rgb]{0.46, 0.05, 0.05}{with} \textcolor[rgb]{0.35, 0.04, 0.04}{whom} \textcolor[rgb]{0.35, 0.04, 0.04}{he} \textcolor[rgb]{0.59, 0.06, 0.06}{talked} \textcolor[rgb]{0.43, 0.05, 0.05}{in} \textcolor[rgb]{0.33, 0.04, 0.04}{the} \textcolor[rgb]{1.00, 0.11, 0.11}{station} \textcolor[rgb]{0.45, 0.05, 0.05}{at} \textcolor[rgb]{0.66, 0.07, 0.07}{fort} \textcolor[rgb]{0.75, 0.08, 0.08}{wayne} \textcolor[rgb]{0.93, 0.10, 0.10}{interested} \textcolor[rgb]{0.37, 0.04, 0.04}{him}

 \\
 
 1820 to 1850 &  \textcolor[rgb]{0.55, 0.06, 0.06}{with} \textcolor[rgb]{0.53, 0.06, 0.06}{whom} \textcolor[rgb]{0.42, 0.05, 0.05}{he} \textcolor[rgb]{0.80, 0.09, 0.09}{talked} \textcolor[rgb]{0.51, 0.06, 0.06}{in} \textcolor[rgb]{0.37, 0.04, 0.04}{the} \textcolor[rgb]{0.89, 0.10, 0.10}{station} \textcolor[rgb]{0.53, 0.06, 0.06}{at} \textcolor[rgb]{0.74, 0.08, 0.08}{fort} \textcolor[rgb]{0.48, 0.05, 0.05}{wayne} \textcolor[rgb]{1.00, 0.11, 0.11}{interested} \textcolor[rgb]{0.33, 0.04, 0.04}{him}

 \\
 
  1850 to 1880 &  \textcolor[rgb]{0.45, 0.05, 0.05}{with} \textcolor[rgb]{0.49, 0.05, 0.05}{whom} \textcolor[rgb]{0.38, 0.04, 0.04}{he} \textcolor[rgb]{0.68, 0.07, 0.07}{talked} \textcolor[rgb]{0.48, 0.05, 0.05}{in} \textcolor[rgb]{0.33, 0.04, 0.04}{the} \textcolor[rgb]{0.70, 0.08, 0.08}{station} \textcolor[rgb]{0.47, 0.05, 0.05}{at} \textcolor[rgb]{0.60, 0.07, 0.07}{fort} \textcolor[rgb]{0.79, 0.09, 0.09}{wayne} \textcolor[rgb]{1.00, 0.11, 0.11}{interested} \textcolor[rgb]{0.35, 0.04, 0.04}{him}

 \\
  
   1880 to 1910 &  \textcolor[rgb]{0.52, 0.06, 0.06}{with} \textcolor[rgb]{0.51, 0.06, 0.06}{whom} \textcolor[rgb]{0.46, 0.05, 0.05}{he} \textcolor[rgb]{0.69, 0.08, 0.08}{talked} \textcolor[rgb]{0.48, 0.05, 0.05}{in} \textcolor[rgb]{0.38, 0.04, 0.04}{the} \textcolor[rgb]{0.52, 0.06, 0.06}{station} \textcolor[rgb]{0.46, 0.05, 0.05}{at} \textcolor[rgb]{0.65, 0.07, 0.07}{fort} \textcolor[rgb]{1.00, 0.11, 0.11}{wayne} \textcolor[rgb]{0.97, 0.11, 0.11}{interested} \textcolor[rgb]{0.33, 0.04, 0.04}{him}

 \\
   
   1910 to 1940 &  \textcolor[rgb]{0.46, 0.05, 0.05}{with} \textcolor[rgb]{0.44, 0.05, 0.05}{whom} \textcolor[rgb]{0.38, 0.04, 0.04}{he} \textcolor[rgb]{0.56, 0.06, 0.06}{talked} \textcolor[rgb]{0.46, 0.05, 0.05}{in} \textcolor[rgb]{0.36, 0.04, 0.04}{the} \textcolor[rgb]{0.44, 0.05, 0.05}{station} \textcolor[rgb]{0.42, 0.05, 0.05}{at} \textcolor[rgb]{0.57, 0.06, 0.06}{fort} \textcolor[rgb]{1.00, 0.11, 0.11}{wayne} \textcolor[rgb]{0.70, 0.08, 0.08}{interested} \textcolor[rgb]{0.33, 0.04, 0.04}{him} \\

          \bottomrule
    \end{tabular}
    \caption{Normalized perplexities for scratch models, lighter red signifies higher surprisal.}
    \label{tab:log_perplexities}
\end{table*}

\subsection{Fluency and Historical Specificity}

\textbf{Perplexity.} We calculate perplexity for the scratch and finetuned models using both the test set drawn from its own historical period and each test set from every other historical slice. We use the former perplexity score to measure model fluency, and the latter scores, which we call "cross-time perplexity" as a measure of historical specificity.

We consider a fluent and well-adapted model to have absolutely low perplexity over the test set drawn from its own time period and relatively high perplexity over those drawn from all other time periods.

\textbf{BabyLM Evaluation Pipeline: BLiMP.} The BabyLM evaluation pipeline provided by \citet{babylm-2024} is a version of EleutherAI's lm-evaluation-harness \citep{eval-harness}, modified to support the evaluation of models trained over a token-limited corpus, in our case the in-vocabulary overlap for all of the time slices. We ensure all test samples are in-vocabulary by filtering to the set of samples where every word appears twice in all model training sets (\emph{maximally filtered}). The pipeline supports evaluation over BLiMP, GLUE \citep{wang-etal-2018-glue} and 
EWoK \citep{ivanova2024elements} of which we solely utilize BLiMP \citep{warstadt-etal-2020-blimp-benchmark}. Concretely, BLiMP tests the model's ability to understand different linguistic phenomena, which we aggregate to measure model fluency. The maximally filtered BLiMP evaluation set retains 25373 of standard BLiMP's 67000 examples. We report the distribution of retained subtasks in Appendix~\ref{sec:blimpdetails}.

\textbf{Novel word sense cloze evaluation set}
We construct this dataset using the Oxford English Dictionary (\emph{OED}), which catalogs English words and their respective word senses. For each word sense the OED provides the year of its first attested usage, as well as example sentences illustrating the word sense in context.
To generate a usable cloze task for next-token prediction models without the ability to follow instructions, the masked words need to be located at the end of the sentences. We select sentences where the word in question appears within the last 10\% of characters. For practicality, we further restrict the dataset to words the OED doesn't consider exceptionally rare, specifically ones appearing once every thousand to a million words (Table~\ref{tab:word_frequency} in Appendix~\ref{sec:clozedetails}). We filter the dataset in the same manner as the BLiMP task, removing samples with uncommon words that have less than two occurrences in any training set.

We perform evaluation by generating the top $k$ one-word model responses to each datapoint. We achieve this using a custom Transformers \citep{wolf-etal-2020-transformers} LogitsProcessor, which redistributes the probability mass of tokens initiating a new word to the EOS token. In combination with probability-based beam search (length penalty set to zero) this method efficiently approximates the top $k$ one word responses. We were unable to find a similar approach in the literature. Example tasks are shown in Table \ref{tab:word_sense}. \\
We use this task to evaluate historical specificity by defining a measure called 'leakage'. Leakage measures how much knowledge a model $m_{(t_0, t_1)}$ modeling a specific time period $(t_0, t_1)$ has of future word senses. An ideal historical model should be familiar with word senses introduced before its knowledge cutoff year, and unfamiliar with those after. Let $\mathrm{Top}_k({m_{(t_0,t_1)}, x)}$ be the set of the model's first $k$ completions for cloze task $x$, and $w^*(x)$ denote the correct answer. For each model we split the cloze tasks into $T_{t_1}$, word senses introduced before or in year $t_1$, and $F_{t_1}$ for senses introduced after. We consider a model to complete a cloze task if the targeted word is within the top 100 answers:
\begin{equation*}
    C_{m(t_0,t_1), k} = \{x: w^*(x) \in Top_{k}({m_{(t_0,t_1)}, x)}\}
\end{equation*}

Leakage is defined as accuracy over future senses:
\begin{equation*}
    l_{m(t_0,t_1)} = \dfrac{|C_{m(t_0,t_1), 100} \cap F_{t_1}|}{|F_{t_1}|}
\end{equation*}
However, we observed that since new word senses don't emerge arbitrarily, stronger language models appear to have higher leakage. Some future cloze tasks can be correctly completed using prior word senses, such as 'silver' in Table \ref{tab:clozeout}. To filter out this effect, we introduce 'recall normalized leakage' which normalizes leakage using the model's pre-cutoff recall:
\begin{equation*}
    r_{m(t_0,t_1)} = \dfrac{|C_{m(t_0,t_1), 100} \cap T_{t_1}|}{|T_{t_1}|}
\end{equation*}

The recall normalized leakage is defined as: 
\begin{equation*}
    RNL_{m(t_0,t_1)} = \dfrac{l_{m(t_0,t_1)}}{r_{m(t_0,t_1)}}
\end{equation*}

More details on sense distribution and evaluation can be found in Appendix \ref{sec:cloze_evaluation}. 

\begin{figure*}[h]
    \centering
    \begin{subfigure}[b]{0.49\linewidth}
        \centering
        \includegraphics[width=\linewidth]{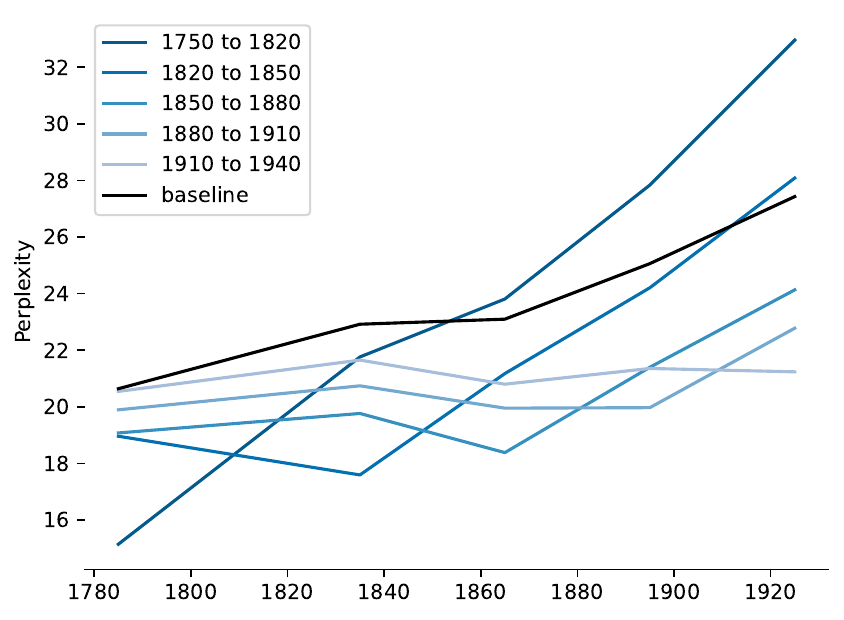}
        \caption{Finetuned models with unadapted Llama3-8B baseline}
        \label{fig:ftp}
    \end{subfigure}
    \hfill
    \begin{subfigure}[b]{0.49\linewidth}
        \centering
        \includegraphics[width=\linewidth]{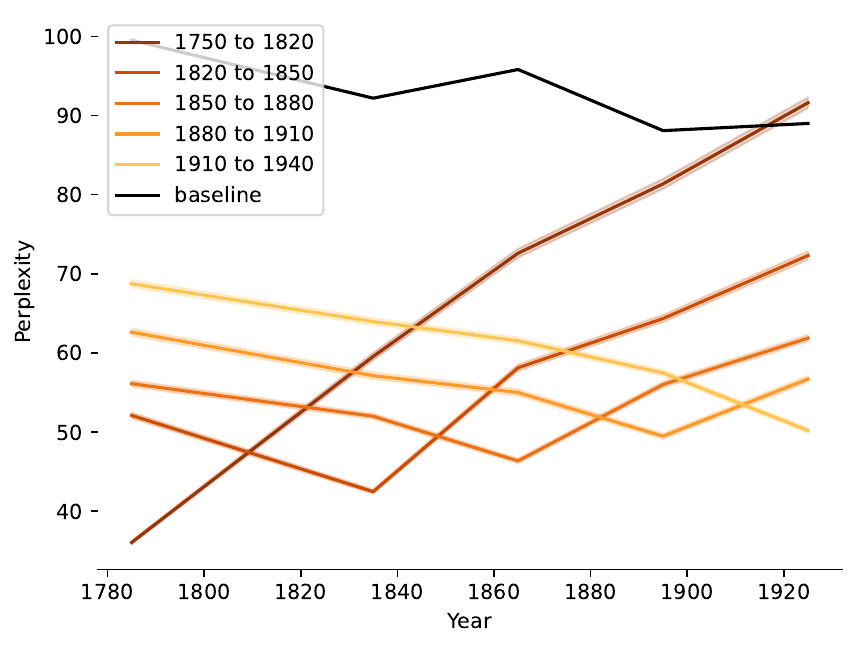}
        \caption{Scratch models with unadapted BabyLlama-2 baseline}
        \label{fig:sctp}
    \end{subfigure}
    \caption{Cross-time perplexities}
    \label{fig:crosstimeperp}
\end{figure*}

\begin{table*}[h]
\centering
\begin{tabular}{llllll}
\toprule
\textbf{Model}     & \textbf{1750-1820} & \textbf{1820-50} & \textbf{1850-80} & \textbf{1880-1910} & \textbf{1910-40} \\
\midrule
Scratch & 0.70$\pm$0.00      & 0.71$\pm$0.00      & 0.73$\pm$0.01      & 0.74$\pm$0.00      & 0.74$\pm$0.00      \\
Finetuned      & 0.80      & 0.81      & 0.83      & 0.84      & 0.84     \\
\bottomrule
\end{tabular}
\caption{Aggregate maximally filtered BLiMP accuracy across all timeslices.}
\label{tab:blimpacc}
\end{table*}

\subsection{Exploratory Analysis}

We perform bottom-up and top down exploratory analysis. To perform bottom-up analysis we contrast the min-max normalized log perplexity of the models over a given sentence. Despite the models having different baseline perplexities, their normalized log perplexity follows a similar trajectory, with the exception of words particularly characteristic (time period-specific) for a model's dataset. This phenomenon is shown in Table \ref{tab:log_perplexities}, where a significant shift can be seen for "station", which lowers in perplexity as the railway system is widely adopted during the 1840s and 50s.
We use this perplexity data to generate candidates for word sense change, motivated by the notion that words whose later sense has not yet emerged should prove more perplexing for earlier models. We also demonstrate how the OED-derived cloze task can be used in a top-down fashion to characterize diachronic change in words known to be historically-sensitive.

\begin{figure*}[tbp]
    \centering
    \begin{subfigure}[b]{0.49\linewidth}
        \centering
        \includegraphics[width=\linewidth]{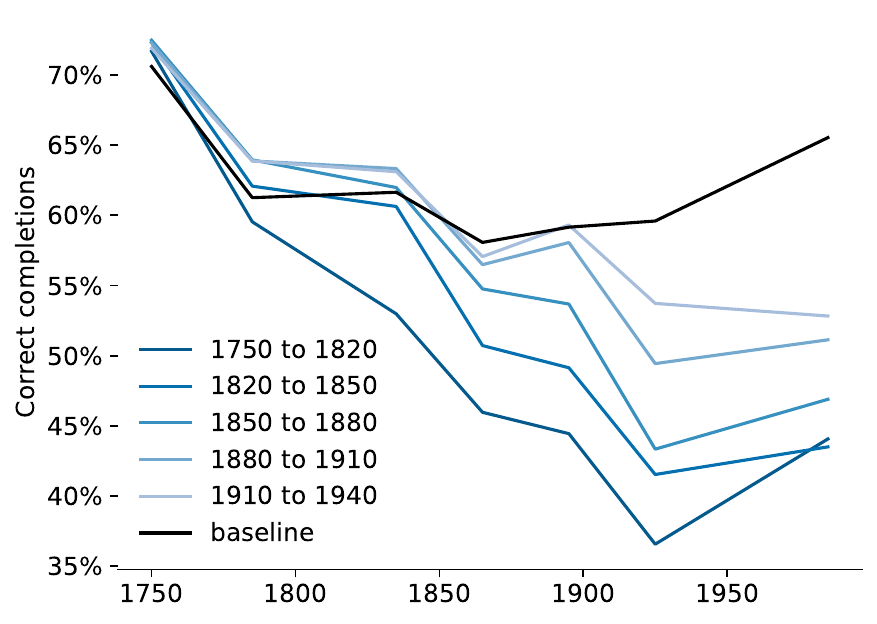}
        \caption{Finetuned models with unadapted Llama3-8B baseline}
        \label{fig:blperp_top100}
    \end{subfigure}
    \hfill
    \begin{subfigure}[b]{0.49\linewidth}
        \centering
        \includegraphics[width=\linewidth]{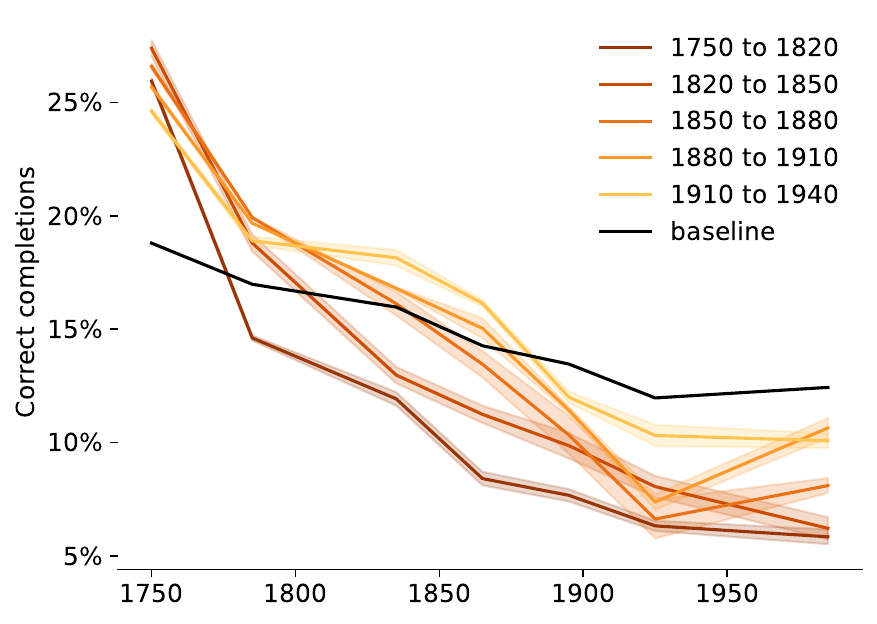}
        \caption{Scratch models with unadapted BabyLlama-2 baseline}
        \label{fig:blperp_top100_baby}
    \end{subfigure}
    \caption{Model performance on the top 100 completion cloze task}
    \label{fig:top100perf}
\end{figure*}

\section{Results and discussion}
\subsection{Fluency and Historical Specificity}
\textbf{The unadapted and finetuned models are fluent, but lack historical specificity.} The unadapted baseline models show slight time-period preferences but are logically a-historical (Figure~\ref{fig:ftp}). The finetuned models have overall low perplexity. Models trained on earlier time slices show a strong preference for data from their respective time slice, whereas those trained on later slices do not show such prominent specialization.

\textbf{The scratch models are less fluent, but are specialized to their historical period.} The scratch models uniformly produce their lowest perplexities when measured against their period's corresponding test set. Cross-evaluation of the models against the other reserved test sets (Figure~\ref{fig:sctp}) yields encouraging signs that the linguistic information captured by the scratch models follows an appropriate historical arc, and that there is little to no information from any of the extraneous time slices. Perplexity is low in the relevant period and increases monotonically both when testing older and newer texts, a sign of period-specific adaptation unique to the scratch models.

\textbf{The scratch models perform reasonably on BLiMP despite generally underperforming the unadapted and finetuned models.} The scratch models earn accuracy scores ranging from 0.70 to 0.74 (Table~\ref{tab:blimpacc}) on the maximally filtered version of the BLiMP evaluation set. For comparison, The two unadapted models earn aggregate scores of 0.74 (BabyLlama-2) and 0.82 (Llama3-8B) on the same set of BLiMP tasks. We expect this slight to moderate underperformance in both cases, as scratch was trained over more specific mix of 10M tokens than unadapted BabyLlama-2 and far fewer tokens than unadapted Llama3-8B. Nonetheless, scratch approaches the general competence of unadapted BabyLlama-2, and meets it in the later slices. The finetuned models consistently outperform the scratch models (Table~\ref{tab:blimpacc}). Beyond verifying that the models are usable models of language, we care about the contrastive differences between them. We note an increase in BLiMP competency over time. Given that BLiMP's test samples are rendered in modern English, this demonstrates that our pretraining approach is potentially sensitive to diachronic grammatical change.

\begin{table*}[bt]
    \centering
    \begin{tabular}{p{4cm} p{4.2cm} c c c}
        \toprule
          \textbf{Sentence}  & \textbf{Definition} & \textbf{Year} & \textbf{Scratch} & \textbf{Finetuned} \\ 
          \midrule
          I'm going to sell my car... No more police towing [it] ..to a car \textbf{pound}. & A place in which vehicles impounded by the police or other authorities are kept... & 1970 & 101$\pm$0.00 & 0 \\ \\
          Hill 
          ... which won three gold and a \textbf{silver}. & Elliptical for silver medal n. & 1960 & 11.67$\pm$5.86 & 0 \\
          \bottomrule
    \end{tabular}
    \caption{Two examples for time slice 1750-1820 with their rank per model. Higher = more historically accurate. Rank 101 indicates the correct word is outside of the top $k=100$. Rank 0 represents the models top choice.}
    \label{tab:clozeout}
\end{table*}

\begin{figure}[h]
    \centering
    \includegraphics[width=\linewidth]{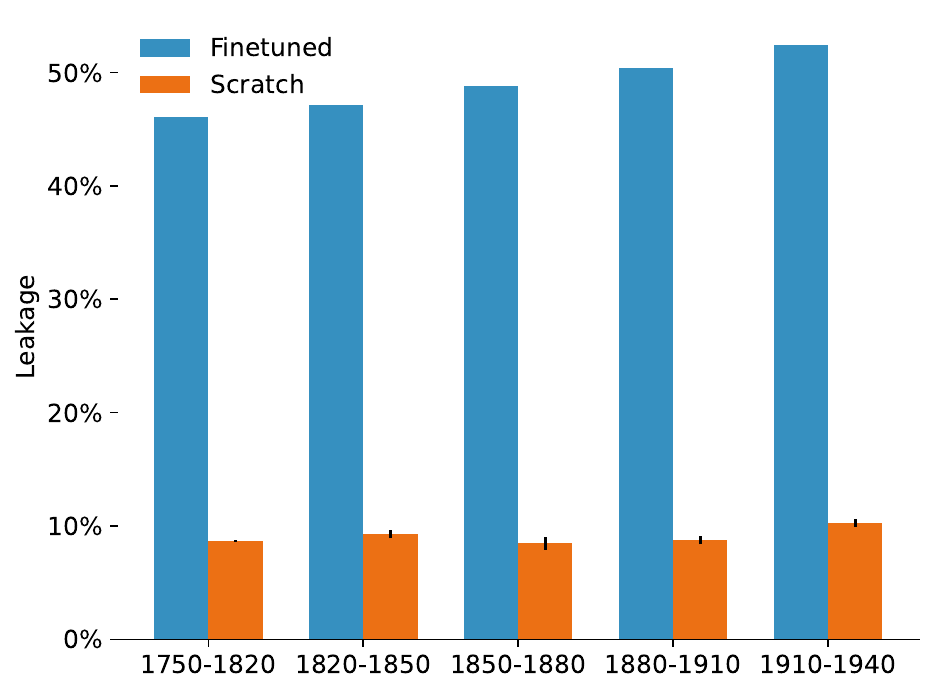}
    \caption{Leakage probability over scratch and finetuned models}
    \label{fig:leakage}
\end{figure}
\begin{figure}[t]
    \centering
    \includegraphics[width=\linewidth]{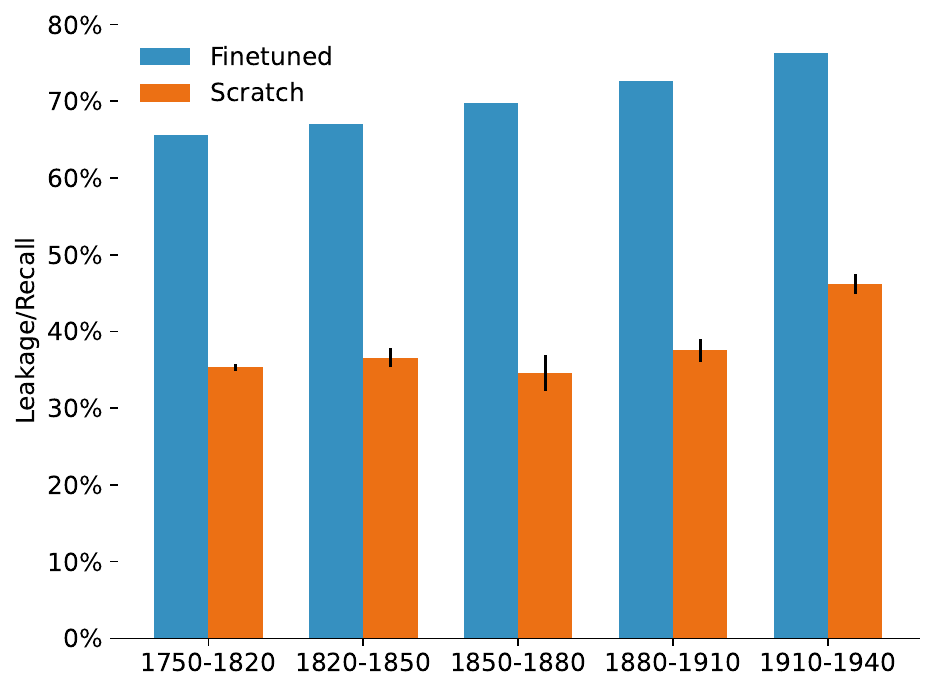}
    \caption{Leakage probability corrected by recall}
    \label{fig:leakage_over_precision}
\end{figure}

\textbf{The scratch models respect diachronic lexical change in a way the finetuned models do not.} While the finetuned models generally "succeed" more in completing a given cloze sentence drawn from OED (Figure \ref{fig:top100perf}), they achieve this in part by integrating information that defies the arrow of time, rendering them less useful for contrasting corpora and studying change. The leakage calculations collected in Figure \ref{fig:leakage} demonstrate that the finetuned models consistently overperform on future senses. Dividing model leakage by recall (Figure \ref{fig:leakage_over_precision}) shows that the finetuned models performance on future time slices remains inappropriately high when correcting for the performance of the scratch models. For details on cloze performance, see Appendix~\ref{sec:cloze_performance}.

Performing error analysis over each slice of both the scratch and finetuned models reinforces the conclusion that finetuned models from early timeslices perform inappropriately well on future cloze tasks, while the scratch models do not. Table~\ref{tab:clozeout} presents two examples of this phenomenon for models trained on the 1750-1820 timeslice. The first example, "pound" is ranked as most likely by the finetuned model and as highly unlikely by the scratch model. This car-centric sense cannot be in the training data for the 1750-1820 slice, as it predates the practical internal combustion engine. Thus, it is probable that the finetuned model is influenced by modern information it ingested in pretraining, resulting in "pound" being ranked as inappropriately likely.

The second example, "silver" as elliptical for "silver medal," is completed by finetuned but also ranked as reasonably likely by scratch. However, this example includes a collocation of "gold," that may steer a model towards a higher probability of "silver." This implies a trend. While the scratch models can perform better than they "should", this likely only occurs when there is a collocated context clue. In contrast, the finetuned model performs out of its bounds on samples like "pound" that lack such clues. This, in combination with the perplexity, leakage and BLiMP evaluations, leads us to conclude that the \textbf{scratch models represent a better tradeoff between fluency and historical accuracy than the finetuned set.}

\begin{table*}[h]
    \centering
    \begin{tabular}{p{4.8cm} p{6.2cm} c}
        \toprule
          \textbf{Sentence}  & \textbf{Definition} & \textbf{Year} 
          \\ 
          \midrule
          They have nowhere to go. This is—how do the Americans say it?—the end of the \textbf{line}. & V. A direction or course of movement. the end of the line ( transferred and figurative ). Cf. the end of the road at end n.. & 1948 
          \\
          \bottomrule
    \end{tabular}
    \caption{The new sense of "line" is accepted by the finetuned (rank \#1) and scratch (\#14) models trained over the 1750-1820 timeslice}
    \label{tab:clozedisc}
\end{table*}  

\subsection{Exploratory analysis}\label{exploratory}

\begin{table*}[h]
    \centering
\begin{tabular}{c c c c c c}
    \toprule
          &  \textbf{1750-1820} & \textbf{1820-1850} & \textbf{1850-1880} & \textbf{1880-1910} & \textbf{1910-1940} \\
          \midrule
Scratch & 12$\pm$4.36                 & 101$\pm$0.00                & 48.67$\pm$15.95                & 78.67$\pm$9.50                & 101$\pm$0.00                \\
Finetuned       & 18                 & 19                 & 11                 & 14                 & 11          \\
            \bottomrule
\end{tabular}
    \caption{Rank of "cholera" completion. Unadapted Llama3-8B ranks it 8, unadapted BabyLlama-2 ranks it 57. Rank 101 indicates it is outside the top $k=100$.} 
    \label{tab:pigs}
\end{table*}

\textbf{The scratch models enable novel hypotheses about lexical changes across our corpus.} One such approach takes advantage of a type of seeming diachronic error that differs from those explored above. Table~\ref{tab:clozedisc} contains an example of this type of error, in the form of a cloze task centering on the phrase "end of the line." Both scratch and finetuned rank the correct completion within the top 20 possibilities. 
However, while we cannot rule out that the finetuned model is achieving accuracy due to its future knowledge (i.e. it has seen this phrase in pretraining), we can do so for its scratch counterpart (a collocation search of the training data reveals that this exact construction is never used). Additionally, unlike the second type of error ("gold...silver") nothing in the context sentence makes ``line'' a likely conclusion by other direct means. 
Thus, the surprising performance of the scratch model is plausibly explained as a \textbf{prefiguration of a construction to come.} The way "line" is used in the 1750-1820 slice of the corpus predicts its ability to be used in this particular construction in the future. Examining the training corpus reveals numerous uses of "line" in hereditary (i.e. "end of one's line") writing (i.e. "the line ended") and martial (i.e. "the British line") contexts, all uses logically associated with the action of "ending." Some uses, especially those related to writing, seem sufficient to support this construction. In a sense, these uses "pave the way" for "end of the line." The restricted training of the scratch models allows them to capture this possible developmental narrative in a way the finetuned models cannot.

Employing the full diachronic battery of scratch models further demonstrates our method's ability to \textbf{detect subtle information about use from the top down}. Table~\ref{tab:pigs} shows the ranking of the correct completion word ("cholera") for each of the models when completing the OED cloze:
\begin{quote}
    The potatoes failed, the pigs were affected with a disease which the people called \textbf{cholera}
\end{quote}

In modern usage, "cholera" denotes the illness associated with the pathogenic bacteria Vibrio cholerae, known historically as "asiatic cholera." Prior to the development of germ theory and the arrival of Cholera vibrio on Western shores, "cholera," often paired with an additional identifier like "morbus" or "infantus," was used to describe a variety of intestinal maladies that shared similar symptoms \citep{rousseau2003coleridge} \citep{barua1992history}.

However, this particular 1837 sense of "cholera" concerns a pig disease that presumably earned its appellation due to its gastric manifestation. The scratch sets' acceptance of this usage follows a distinct trajectory. The earliest model finds the usage most acceptable, the next few models vacillate, and the final (post-germ theory) model rejects it. In contrast, the finetuned models rank it as highly acceptable across the whole period. In doing so, they miss a less-explored facet of cholera's conceptual solidification. Further research reveals that before the term's conceptual solidification around Cholera vibrio, pig "choleras" shared semantic space with their human cousins \citep{cole1961history}. Far from being a minor detail, this helps reveal how "cholera" became less a description of a set symptoms common to the animal kingdom and more a term for a particular human pathogen.

Examining the collocation of "cholera" in the training data of the earliest slice reveals that multiple forms of cholera, including "morbus" and "infantus," share lexical space with descriptions of their symptomatic overlap ("fever", "diarrhea" etc.). In the next slices, the term vacillates between these uses and the emerging understanding of cholera as a specific human disease capable of producing mass illness events, demonstrated by collocations like "epidemic" and "plague." Finally, it settles on its modern meaning (including collocation with "germ") in the last slice. (See supplementary materials). The scratch models capture and characterize this moment of conceptual solidification, hinting at their utility for investigating the details of historical social and knowledge changes from the top-down.

\textbf{Discovering sense trajectories of interest from the bottom-up.}
Contrasting information provided by each scratch timeslice model allows more flexible automated hypotheses generation than the cloze approach utilized above.
By taking this approach, word senses can be said to have trajectories across our corpus periods, as judged by their scratch model acceptability at each slice. For example, one would expect earlier models to be perplexed by the word "car", in the sense of an automobile, and expect the later models would accept it. To track how this shift occurs, we set the 1910-1940 model's normalized per-word perplexity scores as a baseline, and retrieve all words that have a continually decreasing perplexity difference across time. For tractability, we subset this group to those with the largest change in acceptability between the first and last models.

 \begin{figure}[t]
    \centering
    \includegraphics[width=\linewidth]{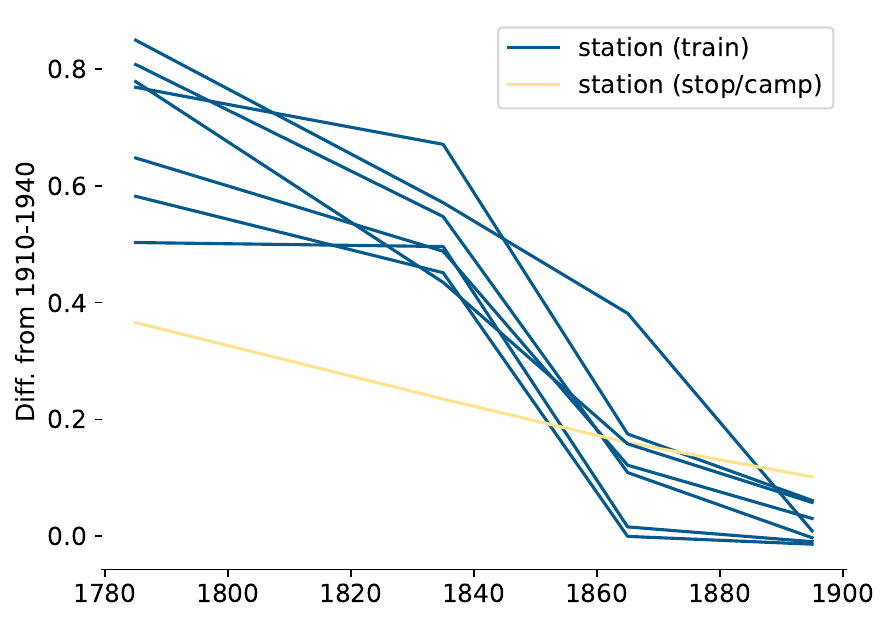}
    \caption{Natural appearances of "station" with a descending probability trajectory and manually labeled for sense. }
    \label{fig:station}
 \end{figure}

This approach captures distinctions in usage over time, and separates synchronically distinct senses of words. Figure~\ref{fig:station} depicts the trajectories for one of these words -- "station." (See the full data in the supplementary materials.) Two senses of the word emerge after applying our filtering approach; one is associated with a railroad station, and one with a stopover or encampment site. While both of these senses becomes more acceptable as time goes on, they follow distinct trajectories. The rail-related sense becomes precipitously more acceptable in the 1820-1850 slice, likely due to the adoption of rail technology during that period. In contrast, the camp/stopover sense begins its trajectory from a place of relative acceptability, and then proceeds to become smoothly more acceptable as time passes.

These observations lead to any number of hypotheses about the interaction between these senses that could be pursued by further means over larger corpora. For example, this information allows the question of whether the already-acceptable usage of "station" as camp or stop grew more acceptable due to the influence of the emerging rail-related sense. We offer some further analytical directions in Appendix~\ref{sec:fac}.

\section{Conclusion and future work}
Our approach leverages efficient pretraining in order to offer a novel, boundary-guaranteed, form of linguistic hypothesis discovery across comparative corpora. We demonstrate that the tradeoff of some model fluency for boundary accuracy pays off by enabling a series of exploratory techniques. While we believe this approach has potential for the specific use case examined above, diachronic change, we also believe that it is extensible, and can be leveraged in a similar way across different corpus divisions and fields. Further work could verify these beliefs by testing our approach's ability to detect linguistic shifts across synchronic boundaries. Additionally, increasing the size of the training corpora and developing historically-situated post-training regimes could lead to increased model capability, allowing for the discovery of knowledge-level hypotheses relevant to disciplines like literary studies and history.

\section{Limitations}
This work has a few primary limitations. The first is selection of data. While we were able to attribute text publication dates with reasonable accuracy, it is not clear that our method will work on less well-documented texts. This had the effect of limiting our training corpora for this study to works like  long-form fiction and nonfiction that are heavily commented on in the historical record. The second is model performance. While our scratch models are reasonably fluent, the tradeoff between performance and historical certainty they traffic in could potentially be further minimized by other efficient training techniques.  Finally, we chose to limit our training data to a single language and modality for this study.

\cready{
\section*{Acknowledgments}
We thank Haim Dubossarsky for feedback concerning the use and evaluation of our modeling strategy. Elisabeth Fittschen acknowledges the support of Hamburglobal, whose stipend partially funded her exchange program.

}

\bibliography{anthology, custom}

\appendix
\section{Additional model and training details}
\label{sec:adddetails}
DoRa adapters of rank 16 were chosen for efficiency purposes. We adopt the hyperparamers used in the original paper (Table~\ref{table:dora_hyperparameters}). 

\begin{table}[h]
    \centering
    \begin{tabular}{l c}
        \toprule
        \textbf{Hyperparameters (DoRA)} & \textbf{LLaMA3-8B} \\
        \midrule
        Rank \( r \) & 16 \\
        \( \alpha \) & 32 \\
        Dropout & 0.05 \\
        Optimizer & AdamW \\
        LR & \( 1 \times 10^{-4} \) \\
        LR Scheduler & Linear \\
        Batch size & 16 \\
        Warmup Steps & 100 \\
        Epochs & 3 \\
        Where & Q, K, V, Up, Down \\
        \bottomrule
    \end{tabular}
    \caption{Hyperparameter configurations of DoRA for LLaMA3-8B.}
    \label{table:dora_hyperparameters}
\end{table}

We finetune using next-token prediction loss on the data slices. We train the models for three epochs, with one training run taking around eight hours on a single A100 GPU, which emits 0.11KG $CO_2$ an hour \citep{lacoste2019quantifying}. We choose the best model checkpoints based on evaluation loss. Notably, models trained on data after 1850 reached optimal performance after a single epoch, while those trained over earlier periods continued to show improvement during the second epoch.

We train the scratch models using the BabyLlama-2 training recipe. We adopt the same Llama-345M model (Table \ref{table:BabyLlama-2_architecture}) and training hyperparameters (Table \ref{table:BabyLlama-2_hyperparameters}) as the original paper. 

\begin{table}[h]
    \centering
    \begin{tabular}{l c}
        \toprule
        \textbf{Hyperparameter} & \textbf{Value} \\
        \midrule
        Learning rate & \( 7 \cdot 10^{-4} \) \\
        Number of epochs & 8 \\
        Batch size & 128 \\
        Weight decay & 5 \\
        \midrule
        Distillation \( \alpha \) & 0.5 \\
        \bottomrule
    \end{tabular}
    \caption{Training and distillation hyperparameters of
BabyLlama-2}
    \label{table:BabyLlama-2_hyperparameters}
\end{table}

\begin{table}[h]
    \centering
    \begin{tabular}{l c}
        \toprule
        \textbf{Hyperparameter} & \textbf{Value} \\
        \midrule
        Vocabulary size & 16,000 \\
        Number of layers & 32 \\
        Number of heads & 15 \\
        Number of KV heads & 5 \\
        Embedding dimension & 960 \\
        Hidden dimension & 2560 \\
        Total parameters & 345M \\
        \bottomrule
    \end{tabular}
    \caption{BabyLlama-2 Model Architecture.}
    \label{table:BabyLlama-2_architecture}
    \
\end{table}

BabyLlama-2 uses a distillation strategy where the logits of two trainer models are used to train a student model. Notably, the teacher and student models are of the same size. We initialize a Byte-Pair-Encoding tokenizer for each time slice and train two teacher models over the training data for eight epochs. We select the model with the best validation score. Training a teacher model took around 32 minutes on a single A100 GPU. From the two teachers, we then distill a student model using the distillation loss after \citet{hinton2015distilling}, with 
$
L = \alpha L_{CE} + (1 - \alpha) L_{KL}
$
This loss is made up in equal parts of the normal next token prediction loss and the loss over the soft trainer logits. We train the student over eight epochs. Training the student model took 3 hours and 20 minutes on an A100.

\section{Attribution pipeline details}
\label{sec:wdad}
We extract from WD all entities with an occupation of "author" or "writer" that also have birth dates that fall within this range. We further constrain this subset by filtering it to only include authors WD indicates were known to write in English.

We then fuzzily match this set of authors to those in PGC. The first pass uses Levenshtein distance matching with a predefined threshold in combination with any extractable birth and death information to match PGC authors to the list sourced from WD. The optional second pass uses only fuzzy string matching with a stricter predefined threshold, and matches any remaining PGC authors to an author from WD. This second pass allows for the inclusion of authors without WD-provided date information, compensating for the further loss in certainty with tighter regulation of name similarity. The result of this stage is a mapping between WD authors and PGC authors with an associated list of their works found in PG.

To validate open source and propriety LLM performance on work-date attribution we manually annotated a sample (n=1054) of known-author works with their date of writing using publication information sourced from internet repositories like the HathiTrust collection (This material is available in the supplement). We then used one open weight model (Llama3.3-70B quantized to 4 bits) and two proprietary models (GPT-4, GPT-4o) to zero-shot attribute the dates of works using the following prompt:

\begin{quote}
    When was the work \{\} by \{\} written? Answer just with the year.
\end{quote}

Where the first \{\} was replaced with the work title and the second \{\} by the work author. We then evaluated performance with a tolerance of +/- 1 year to account for the historically common practice of assigning publication date to copyright year. Noting systematic error in the results provided by the best performing model at this stage (GPT-4o) we collected the set of erroneously attributed texts produced by this model and undertook another round of hand annotation on this set, spending additional effort to source historical materials (publishing industry trade journals, library records) that could disambiguate questionable attributions or provide evidence of earlier publications not in the digitized record. We then re-evaluated the models with tolerances of +/- 1 and 10 years, allowing a match to either date attribution to be acceptable. Additionally, we evaluated the models after disqualifying scores with extreme difference (+/- 50 years) from their ground score, to assess the impact of having a more certain source of information (say, author birth and death dates) that pre-restricts correct answers to a tighter range. Table~\ref{tab:attribperf} shows that while the closed-source models perform the best under these conditions, the open source model performs well enough to serve as a first point of departure.

\begin{table}[h]
    \centering
    \begin{tabular}{lllll}
    \toprule
                & \textbf{+/-1} & \textbf{+/-10} & \textbf{DQ +/-1} & \textbf{DQ +/-10} \\
    \midrule
    Llama3.3 & 0.63          & 0.81           & 0.70             & 0.88              \\
    GPT-4       & 0.74          & 0.89           & 0.87             & 0.99              \\
    GPT-4o      & 0.82          & 0.84           & 0.96             & 0.94 \\
    \bottomrule
    \end{tabular}
    \caption{Performance on work date attribution per LLM. +/- indicates year delta tolerance threshold, DQ indicates that extreme variations from the ground scores (+/-50) were not considered}
    \label{tab:attribperf}
\end{table}

Notably, this approach is flexible -- broader diachronic slices justify tolerating more variance. 

\section{Cloze evaluation set details}
\label{sec:cloze_evaluation}
\begin{figure}[h]
    \centering
    \includegraphics[width=0.48\textwidth]{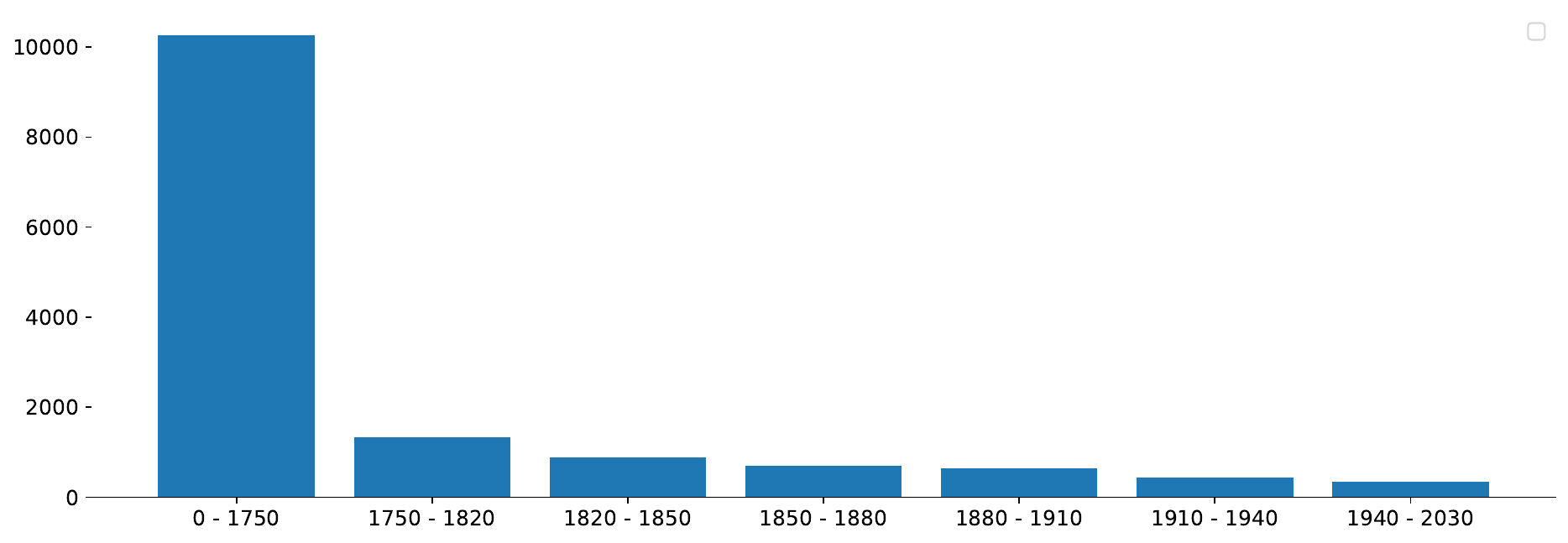}
    \caption{Count of cloze tasks for per time slice for the set filtered for our data (14.6 thousand examples).}
    \label{fig:cloze_distribution}
\end{figure}

The cloze evaluation set contains 50.4 thousand examples. 14.6 thousand examples remain after filtering, a large portion of which is of old english origin as can be seen in Figure \ref{fig:cloze_distribution}. Evaluation was performed over the top 100 word completion task. If the word appeared within the top 100 words (case insensitive) the completion was considered successful. For evaluation the senses were grouped by time slice. In Figures \ref{fig:top100perf} and \ref{fig:mrr_res}, each model was evaluated over each time slice. In the leakage reports (Figures \ref{fig:leakage} and \ref{fig:leakage_over_precision}) the model performance was contrasted between the senses created before and after the models respective training cutoff.  

\label{sec:clozedetails}
\begin{table}[h]
    \centering
    \begin{tabular}{c c c}
        \toprule
        \textbf{Band} & \textbf{Freq./mil.}& \textbf{\% in OED} \\
        \midrule
        8 & $>$1,000 & 0.02\% \\
        7 & 100 -- 1,000 & 0.18\% \\
        6 & 10 -- 100 & 1\% \\
        5 & 1 -- 10 & 4\% \\
        4 & 0.1 -- 1 & 11\% \\
        3 & 0.01 -- 0.1 & 20\% \\
        2 & $<$0.01 & 45\% \\
        1 & -- & 18\% \\
        \bottomrule
    \end{tabular}
    \caption{Word Frequency Bands and their respective counts per a million words and the percentage of non-obsolete OED entries}
    \label{tab:word_frequency}
\end{table}
\newpage

\section{BLIMP evaluation set details}
\label{sec:blimpdetails}
The BLIMP task was filtered from the training data across all time slices, resulting in an uneven distribution over BLIMP subtasks, an overview of this can be seen in Table \ref{tab:blimp_subtask_overview}.

\begin{table}[t]
\centering
\small
\begin{tabular}{lrr}
\hline
\textbf{Task} & \textbf{Filtered} & \textbf{Original} \\
\hline
blimp\_principle\_A\_domain\_2 & 322 & 1000 \\
blimp\_ellipsis\_n\_bar\_2 & 199 & 1000 \\
blimp\_tough\_vs\_raising\_1 & 501 & 1000 \\
blimp\_animate\_subject\_passive & 334 & 1000 \\
blimp\_irregular\_past\_participle\_adjectives & 658 & 1000 \\
blimp\_determiner\_noun\_agreement\_2 & 457 & 1000 \\
blimp\_wh\_vs\_that\_with\_gap\_long\_distance & 364 & 1000 \\
blimp\_sentential\_negation\_npi\_licensor\_present & 478 & 1000 \\
blimp\_wh\_vs\_that\_with\_gap & 486 & 1000 \\
blimp\_determiner\_noun\_agreement\_with\_adj\_irregular\_2 & 390 & 1000 \\
blimp\_irregular\_past\_participle\_verbs & 522 & 1000 \\
blimp\_coordinate\_structure\_constraint\_object\_extraction & 377 & 1000 \\
blimp\_anaphor\_number\_agreement & 603 & 1000 \\
blimp\_wh\_questions\_subject\_gap & 268 & 1000 \\
blimp\_principle\_A\_domain\_1 & 346 & 1000 \\
blimp\_drop\_argument & 468 & 1000 \\
blimp\_determiner\_noun\_agreement\_with\_adjective\_1 & 438 & 1000 \\
blimp\_complex\_NP\_island & 240 & 1000 \\
blimp\_principle\_A\_domain\_3 & 304 & 1000 \\
blimp\_matrix\_question\_npi\_licensor\_present & 486 & 1000 \\
blimp\_left\_branch\_island\_simple\_question & 467 & 1000 \\
blimp\_expletive\_it\_object\_raising & 172 & 1000 \\
blimp\_wh\_questions\_subject\_gap\_long\_distance & 166 & 1000 \\
blimp\_causative & 287 & 1000 \\
blimp\_determiner\_noun\_agreement\_with\_adj\_irregular\_1 & 300 & 1000 \\
blimp\_wh\_questions\_object\_gap & 248 & 1000 \\
blimp\_regular\_plural\_subject\_verb\_agreement\_1 & 314 & 1000 \\
blimp\_determiner\_noun\_agreement\_with\_adj\_2 & 423 & 1000 \\
blimp\_existential\_there\_object\_raising & 193 & 1000 \\
blimp\_inchoative & 506 & 1000 \\
blimp\_superlative\_quantifiers\_1 & 616 & 1000 \\
blimp\_anaphor\_gender\_agreement & 516 & 1000 \\
blimp\_tough\_vs\_raising\_2 & 333 & 1000 \\
blimp\_regular\_plural\_subject\_verb\_agreement\_2 & 511 & 1000 \\
blimp\_determiner\_noun\_agreement\_irregular\_2 & 391 & 1000 \\
blimp\_principle\_A\_case\_2 & 369 & 1000 \\
blimp\_superlative\_quantifiers\_2 & 614 & 1000 \\
blimp\_existential\_there\_subject\_raising & 398 & 1000 \\
blimp\_npi\_present\_2 & 480 & 1000 \\
blimp\_distractor\_agreement\_relational\_noun & 237 & 1000 \\
blimp\_wh\_island & 487 & 1000 \\
blimp\_irregular\_plural\_subject\_verb\_agreement\_1 & 289 & 1000 \\
blimp\_transitive & 263 & 1000 \\
blimp\_sentential\_subject\_island & 372 & 1000 \\
blimp\_principle\_A\_c\_command & 427 & 1000 \\
blimp\_principle\_A\_case\_1 & 333 & 1000 \\
blimp\_existential\_there\_quantifiers\_1 & 410 & 1000 \\
blimp\_principle\_A\_reconstruction & 571 & 1000 \\
blimp\_existential\_there\_quantifiers\_2 & 367 & 1000 \\
blimp\_wh\_vs\_that\_no\_gap\_long\_distance & 217 & 1000 \\
blimp\_adjunct\_island & 305 & 1000 \\
blimp\_animate\_subject\_trans & 269 & 1000 \\
blimp\_distractor\_agreement\_relative\_clause & 291 & 1000 \\
blimp\_coordinate\_structure\_constraint\_complex\_left\_branch & 266 & 1000 \\
blimp\_wh\_vs\_that\_no\_gap & 290 & 1000 \\
blimp\_only\_npi\_scope & 272 & 1000 \\
blimp\_irregular\_plural\_subject\_verb\_agreement\_2 & 503 & 1000 \\
blimp\_passive\_1 & 235 & 1000 \\
blimp\_determiner\_noun\_agreement\_1 & 427 & 1000 \\
blimp\_passive\_2 & 381 & 1000 \\
blimp\_only\_npi\_licensor\_present & 447 & 1000 \\
blimp\_npi\_present\_1 & 470 & 1000 \\
blimp\_ellipsis\_n\_bar\_1 & 214 & 1000 \\
blimp\_left\_branch\_island\_echo\_question & 439 & 1000 \\
blimp\_determiner\_noun\_agreement\_irregular\_1 & 283 & 1000 \\
blimp\_sentential\_negation\_npi\_scope & 269 & 1000 \\
blimp\_intransitive & 494 & 1000 \\
\hline
\textbf{Total} & \textbf{25373} & \textbf{67000} \\
\hline
\end{tabular}
\caption{Overview BLIMP filtered.}
\label{tab:blimp_subtask_overview}
\end{table}

\section{Performance of DoRA finetuned pretrained BabyLlama-2 (small)}
\label{sec:smallperf}
\begin{figure*}[h]
    \centering
    \begin{subfigure}[b]{0.49\linewidth}
        \centering
        \includegraphics[width=\linewidth]{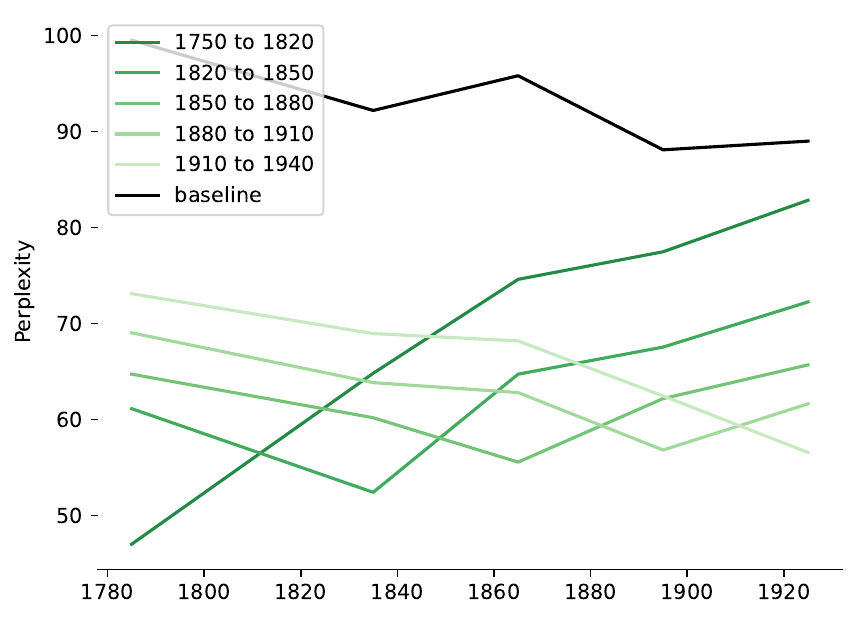}
        \caption{Perplexity of small finetuned models with baseline.}
    \end{subfigure}
    \hfill
    \begin{subfigure}[b]{0.49\linewidth}
        \centering
        \includegraphics[width=\linewidth]{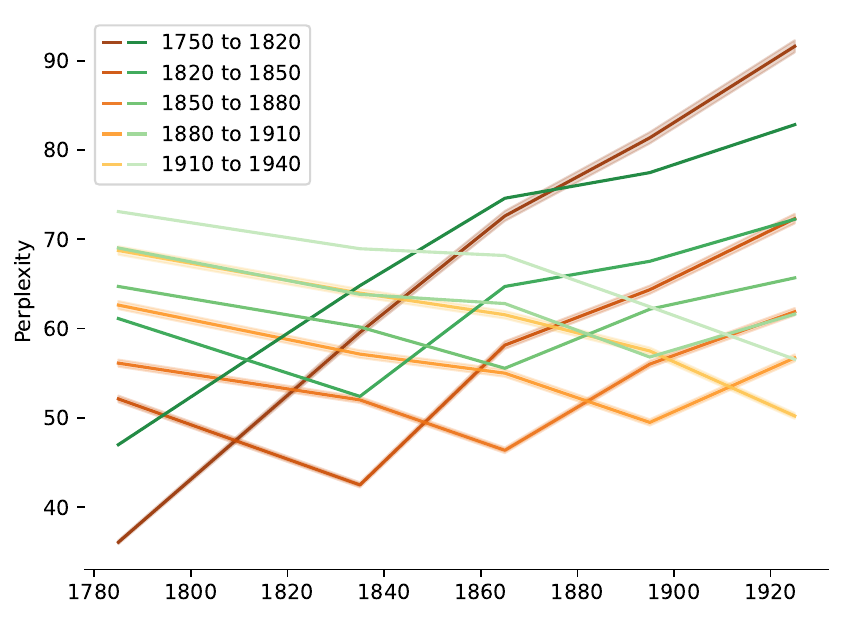}
        \caption{Small finetuned model compared to scratch.}
    \end{subfigure}
    \caption{Cross-time perplexity comparison of small and scratch}
    \label{fig:crosstimeperp_special}
\end{figure*}

\begin{figure*}[h]
    \centering
    \begin{subfigure}[b]{0.49\linewidth}
        \centering
        \includegraphics[width=\linewidth]{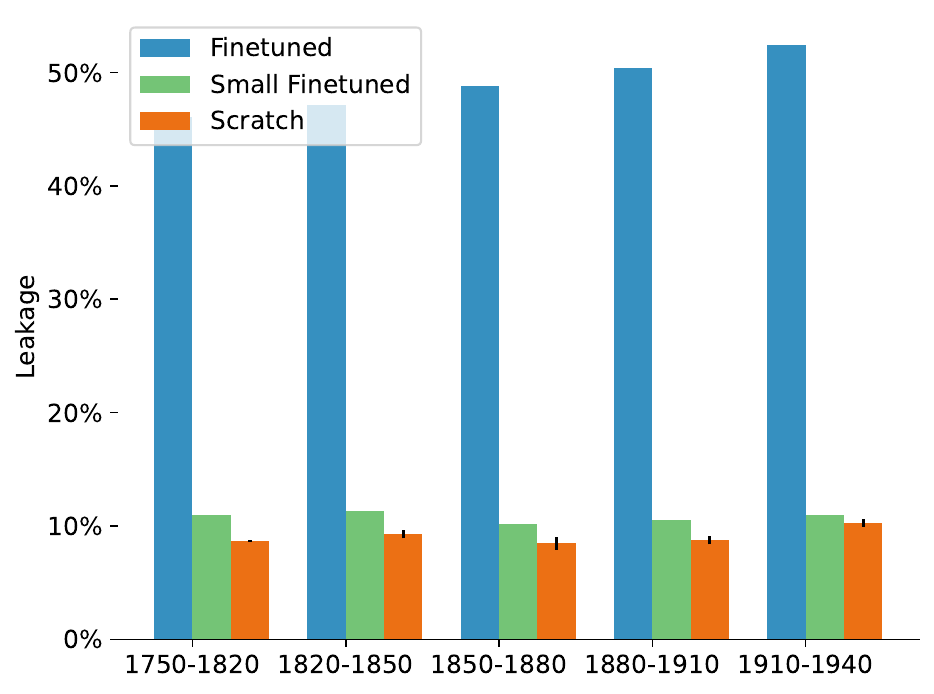}
        \caption{Leakage probability over scratch and finetuned models.}
    \end{subfigure}
    \hfill
    \begin{subfigure}[b]{0.49\linewidth}
        \centering
        \includegraphics[width=\linewidth]{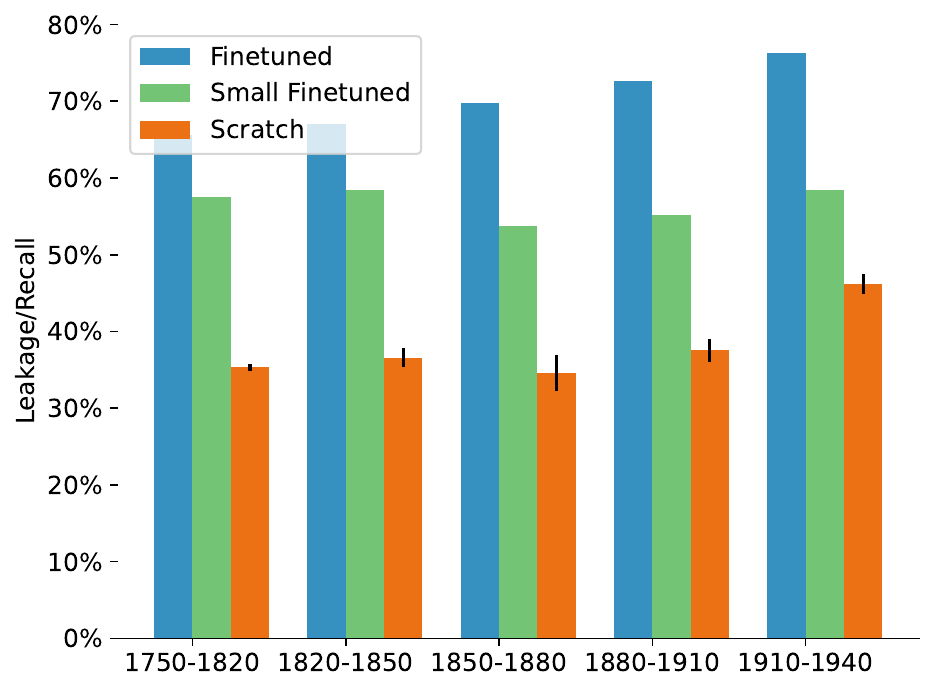}
        \caption{Leakage probability corrected by recall.}
    \end{subfigure}
    \caption{Leakage charts including the small finetuned models.}
    \label{fig:special_leakage}
    
\end{figure*}

As an additional point of comparison, we also finetuned the distribution checkpoint of BabyLlama-2 using the data from our historical timeslices (\textit{'finetuned small'}). This checkpoint of BabyLlama-2, provided by the authors of the BabyLlama-2 paper, was pretrained on the BabyLM mixture of 10 million tokens for 8 epochs. While this comparison is not a perfect one, as the \textit{'finetuned small'} model has prior access to more linguistic information from its pretraining, the resulting small models under-perform the approaches explored in the main body of the paper (Figure~\ref{fig:crosstimeperp_special}). They also suffer from more leakage than scratch (Figure~\ref{fig:special_leakage}).

\newpage

\section{Additional model performance information}
\begin{figure*}[h]
    \centering  
    \begin{subfigure}[b]{0.49\linewidth}
        \centering
        \includegraphics[width=\linewidth]{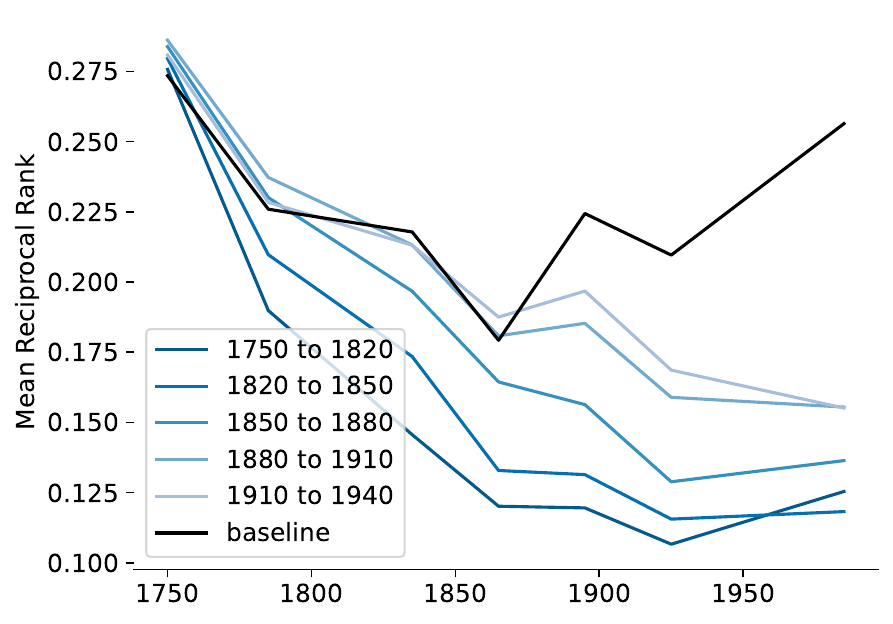}
        \caption{Finetuned MRR with unadapted Llama3 baseline}
        \label{fig:blperp_mrr}
    \end{subfigure}
    \hfill
    \begin{subfigure}[b]{0.49\linewidth}
        \centering
        \includegraphics[width=\linewidth]{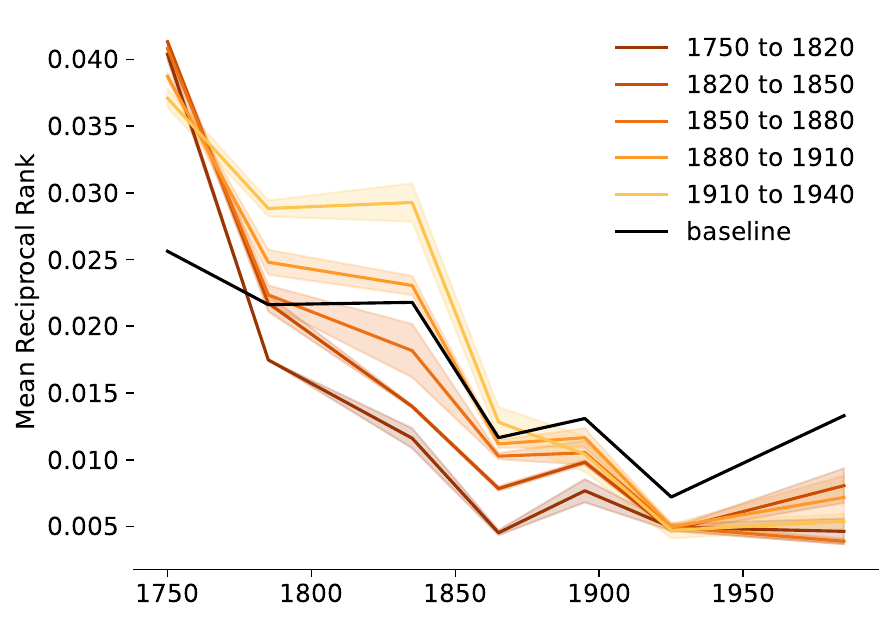}
        \caption{Scratch MRR with unadapted BabyLlama-2 baseline}
        \label{fig:blperp_mrr_baby}
    \end{subfigure}
    \caption{MRR performance on the cloze task}
    \label{fig:mrr_res}
\end{figure*}

We include an overview of mean reciprocal rank over time, for a more detailed insight into model performance (Figure \ref{fig:mrr_res}). 
\label{sec:cloze_performance}
\label{sec:ampi}

\section{Further analytical commentary}
\label{sec:fac}
\begin{figure*}[htbp]
    \centering
    \includegraphics[width=\textwidth]{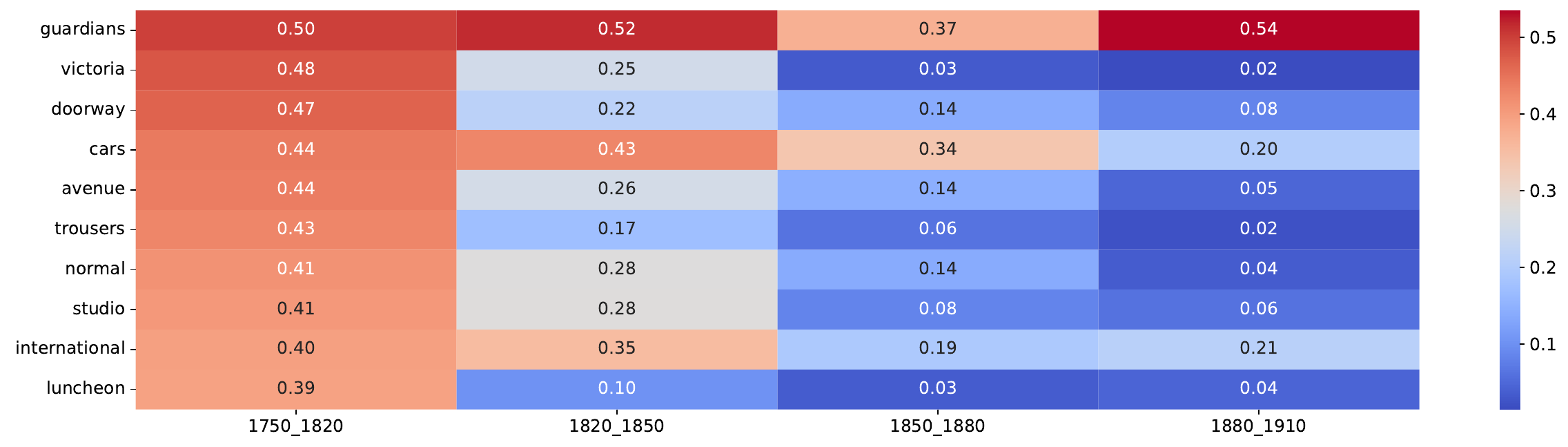}
    \caption{Cumulative perplexity results.}
    \label{fig:cumulative_results}
\end{figure*}

In a second, more cumulative analysis (Figure \ref{fig:cumulative_results}), words with consistently high perplexity differences were highlighted. The underlying reason for these variations is varied. Some words show semantic shifts, such as “car” (automobile) “plane” (airplane) and “inspector” (detective), while others are a part of novel word combinations, which had gained popularity such as “skirt” in the context of “hobble skirt” or “Victoria” in the context “Queen Victoria”. While these insights cannot be pinpointed to a single phenomenon, they offer valuable insights into the training corpora.

\end{document}